\documentclass[conference]{IEEEtran}
\usepackage{times}

\usepackage[numbers]{natbib}
\usepackage{multicol}
\usepackage[usenames,dvipsnames,table]{xcolor}
\usepackage[bookmarks=true]{hyperref}
\hypersetup{
    colorlinks=true,
    linkcolor=orange,
    filecolor=magenta,      
    urlcolor=orange,
    citecolor=orange,
}
\usepackage{xspace}
\usepackage{amsmath}
\usepackage{amsfonts}
\usepackage{dsfont}
\usepackage[capitalize]{cleveref}
\usepackage{pdfx}
\usepackage{graphicx}
\usepackage{multirow}
\usepackage{float}

\newcommand{\E}{\mathbb{E}}
\newcommand{\Ls}{\mathcal{L}}

\newcommand{\D}{\mathcal{D}}
\newcommand{\kl}{D_\text{KL}}

\newcommand{\fullname}[0]{Demonstration Information Estimation\xspace}
\newcommand{\abv}[0]{DemInf\xspace}

\begin{document}

\title{Robot Data Curation \\ with Mutual Information Estimators}


\author{
\IEEEauthorblockN{
Joey Hejna$^{1, 2, *}$,
Suvir Mirchandani$^2$,
Ashwin Balakrishna$^1$, 
Annie Xie$^1$,
Ayzaan Wahid$^1$,
Jonathan Tompson$^1$,
}
\IEEEauthorblockN{
Pannag Sanketi$^1$,
Dhruv Shah$^1$,
Coline Devin$^{1,}$\textsuperscript{\textdagger},
Dorsa Sadigh$^{1,}$\textsuperscript{\textdagger}
}

\IEEEauthorblockA{$^1$Google DeepMind Robotics, $^2$Stanford University}
}

\newcommand\blfootnote[1]{
  \begingroup
  \renewcommand\thefootnote{}\footnote{#1}
  \addtocounter{footnote}{-1}
  \endgroup
}

\maketitle

\begin{abstract}
The performance of imitation learning policies often hinges on the datasets with which they are trained. Consequently, investment in data collection for robotics has grown across both industrial and academic labs. However, despite the marked increase in the quantity of demonstrations collected, little work has sought to assess the quality of said data despite mounting evidence of its importance in other areas such as vision and language. In this work, we take a critical step towards addressing the data quality in robotics. Given a dataset of demonstrations, we aim to estimate the relative quality of individual demonstrations in terms of both action diversity and predictability. To do so, we estimate the average contribution of a trajectory towards the mutual information between states and actions in the entire dataset, which captures both the entropy of the marginal action distribution and the state-conditioned action entropy. Though commonly used mutual information estimators require vast amounts of data often beyond the scale available in robotics, we introduce a novel technique based on $k$-nearest neighbor estimates of mutual information on top of simple VAE embeddings of states and actions. Empirically, we demonstrate that our approach is able to partition demonstration datasets by quality according to human expert scores across a diverse set of benchmarks spanning simulation and real world environments. Moreover, training policies based on data filtered by our method leads to a 5-10\% improvement in RoboMimic and better performance on real ALOHA and Franka setups. \blfootnote{$^*$ Work done as a Student Researcher at Google DeepMind} \blfootnote{\textsuperscript{\textdagger} Equal advising.} \blfootnote{Videos and code at \url{https://jhejna.github.io/demonstration-info}.}

\end{abstract}


\section{Introduction}

Supervised learning via maximum likelihood estimation, i.e., attempting to reproduce the training distribution, underpins several recent advancements in deep learning. 
Due to the broad availability of high-quality data on the internet, models in vision \citep{radford2021learning} and language \citep{radford2019language} have continued to improve through supervised learning on larger and larger datasets \citep{kaplan2020scaling, zhai2022scaling}.
The observed trend of more data leading to more performance has inspired parts of robot learning community, spurring increased investment in data collection across both academia and industry \citep{openx, rh20t, droid, bridge} in hopes of training better imitation learning policies, often with similar maximum likelihood objectives \citep{octo,rt1}. However, MLE-based approaches benefit most from high-quality data, and as we have seen in vision and language, not all data is equal \citep{xu2024demystifying}. In other words, we should expect the performance of a large behavior cloning policies to mirror the quality of the collected data, and we may not be gathering the most optimal data. As an example, the recent DROID dataset \citep{droid} contains 76K demonstrations collected across 13 institutions. Despite being one of the largest and most diverse datasets in robotics, the DROID dataset was significantly down-weighted when training OpenVLA \citep{kim2024openvla} as it was found to hurt performance rather than help. While many hypotheses surrounding this exist, e.g., insufficient operator supervision leading to an excessively broad data distribution, one conclusion we can draw is that we should pay more attention to data \emph{quality}, not just quantity. This is particularly important in robotics, as every demonstration requires labor, time, and capital. 

Even though data quality is critical to the performance of imitation learning algorithms, little work has sought to measure it \citep{belkhale2024data} and unfortunately, techniques used for data curation in vision and language do not transfer well to robotics. For example, \textit{n}-gram classifiers have been extremely effective for web text \citep{albalak2024survey} but are unable to handle high-dimensional continuous states and actions. Pre-trained models have been used to curate vision datasets \citep{schuhmann2022laion}, but are incapable of reasoning about actions. In contrast, we believe metrics for imitation learning should be able to measure the relative predictability of the state-action distribution directly, which affects how well a policy is able to fit the expert \citep{belkhale2024data}. In this work we explore how this desiderata can be captured by the mutual information between states and actions. 

Mutual information, or the bits of information learned about one random variable by observing another, precisely measures the difference between the marginal entropy of one variable and the conditional entropy of another. In the context of states and actions, this means that high mutual information encourages a large diversity of actions (action entropy) but a predictable action distribution (low entropy of actions given a particular state).  We thus propose using mutual information as a desirable metric for measuring data quality in imitation learning. 

Unfortunately, estimating mutual information is particularly hard especially with low amounts of data. Common estimation techniques like InfoNCE \citep{oord2018representation} used for models such as CLIP \citep{radford2021learning} often require millions of data points from the same distribution. In robotics, we often do not have access to data at a similar scale due to the difficulty and cost of collection. Moreover, even if we assume access to more data, existing large robot datasets such as OpenX Embodiment \citep{openx} have sporadic support across a few highly varied environments likely containing little to no overlap with new data collected across different labs, platforms, and tasks. To address this problem, we introduce \fullname or \abv for short. We design \abv to work across both low- and high-data regimes in robotics. To do so, \abv first learns a structured low-dimensional representation of the state and action space using variational autoencoders. Then, \abv leverages mutual information estimators based on $k$-nearest neighbors to estimate the quality of state and action chunk pairs. Critically we found these non-parametric estimators to be more stable with datasets of 50-300 demonstrations commonly used in robotics. Finally, we average mutual information estimates across time to identify the highest and lowest quality demonstrations.

When applying \abv to a number of different robot platforms and environments, we find that it is able to consistently partition high- and low-quality data as scored by expert human annotators, outperforming both contemporary baselines and alternative mutual information estimators like InfoNCE \citep{oord2018representation}. Furthermore, using \abv to subsample demonstration data, we are able to attain higher performing imitation learning policies across the RoboMimic \citep{robomimic} benchmark and real ALOHA policies on RoboCrowd \citep{mirchandani2024robocrowd}.

\section{Related Work}
\label{sec:related_work}
As deep learning models have continued to scale, data quality estimation has become an area of increasing interest. Here we review works most relevant to our approach.

\textbf{Data Quality in Vision and Language.} Data quality has most often been studied in the vision and language domains, where modern training pipelines often include multiple steps of quality estimation and de-duplication \citep{albalak2024survey, together2023redpajama, penedo2024fineweb}. For text data, this often consists of simple $n$-gram classifiers, or meta-data filtering, which have been shown to have a large impact on performance \citep{xu2024demystifying}. Other more advanced techniques use unsupervised clustering  \citep{vo2024automatic, NEURIPS2023_a8f8cbd7, abbas2023semdedup, birodkar2019semantic}, most commonly for de-duplication and balancing across clusters. Though these methods improve the diversity of large datasets, they work mostly at scale and independent of label (or in our case action) quality. Methods in group mixing have been shown to increase learning efficiency by improving the dataset's distributional properties, but do so only at the coarse group level \citep{chen2024aioli, chen2024skill, doremi, fandoge}. This is problematic in robotics as we often are actively collecting data, and want to assess trajectories individually. Most related to our approach are techniques based on pre-trained models such as CLIP \citep{fang2024data}. Though these methods do not explicitly make the connection, contrastive models precisely estimate a bound on mutual information \citep{ma2018noise, oord2018representation}. Due to the data requirements of contrastive learning, such techniques rely on large pre-trained models, e.g., CLIP \citep{radford2021learning}, as priors for curation \citep{fang2024data}. Unfortunately, such priors are useless for estimating the mutual information between states and actions. Moreover, training a similar contrastive model from scratch requires hundreds of millions of training examples \citep{xu2024demystifying}. This is rather unrealistic for the current behavior cloning paradigm where even the largest datasets have less than 100,000 demonstrations \citep{droid}, and we do not have access to strong pre-trained action priors.

\textbf{Data Quality in Robotics.} Orthogonal to us, several works have focused on increasing the size of robotics datasets, through the development of tools \citep{pmlr-v155-young21a, chi2024universal}, human teleoperation \citep{droid,openx,jang2022bc,dasari2019robonet,sharma2018multiple,mandlekar2018roboturk,pinto2016supersizing, bharadhwaj2024roboagent, robocasa2024}, or automatic data augmentation  \citep{ha2023scalingup,mandlekar2023mimicgen, belkhale2023hydra, mandi2022cacti}, with the aim of training large-scale robot policies \citep{rt1, rt2, levine2018learning, octo}. Through this process, data quality in the context of robot learning has come into question, but largely through the lens of inter-demonstration compositional generalization to new objects or scenes \citep{burns2023makes, xie2023decomposing,gao2024efficient, lin2024datascalinglawsimitation}. Unlike our work, such approaches do not consider intra-demonstration transition quality, e.g., how good the action labels are which can ultimately determine the performance of imitation learning methods. Other works that consider action quality do so at an extremely coarse level. ReMix \citep{hejna2024remix} learns group weights over large robot datasets using robust optimization. Such dataset mixing approaches require datasets to be partitioned into groups a priori, and are thus unable to determine the quality of individual demonstrations. Retrieval methods \citep{nasiriany2022sailor,du2023behavior,lin2024flowretrieval} use a target dataset to retrieve state-action pairs from unstructured data, but do not explicitly measure data quality, only similarity. Perhaps most related to our work, \citet{kuhar2023learning} directly estimate the quality of individual demonstrations using a latent space from temporal contrastive learning. However, to actually produce quality estimates they assume access to a dataset of human quality labels. Moreover, the choice of temporal contrastive learning means that the learnability of actions is not explicitly considered. \abv on the other hand is completely unsupervised, and thus can be applied to broad amounts of robot data without any hand annotation.

\textbf{Mutual Information Estimation.} Mutual information estimation has been a long studied problem in both statistics and deep learning \citep{poole2019variational}. Direct mutual information objectives like InfoNCE \citep{oord2018representation} are often used for representation learning in vision \citep{chen2020simple} and language \citep{zhang2020unsupervised}. Other works have used the dual formulation \citep{belghazi2018mutual}. Unfortunately, these parametric methods techniques often require on the order of a million samples for accurate estimation, but having access to this scale of data is rather uncommon when trying to measure data quality for imitation learning in a specific environment. Instead, \abv uses non-parametric estimators based $k$-nearest-neighbors \citep{kozachenko1987sample, singh2003nearest}, specifically the KSG estimator \citep{kraskov2004estimating,gao2018demystifying}. Prior works have used $k$-nn estimators in unsupervised RL, but do so for maximizing state entropy \citep{liu2021behavior, kim2024accelerating}, not mutual information or data quality.

\section{Preliminaries}
\label{sec:preliminaries}

\subsection{Imitation Learning}
Broadly, the objective of imitation learning is to learn a policy $\pi_\theta: \mathcal{S} \rightarrow \mathcal{A}$ parameterized by $\theta$ that is able to effectively reproduce the behavior of an expert $\pi_E$ within an environment with state space $\mathcal{S}$, action space $\mathcal{A}$ and horizon $T$.
Typically, we measure the similarity between the policy and expert using a divergence between their state visitation distributions:
\begin{equation}
    \min_\theta \kl\left(\rho_{\pi_\theta} || \rho_{\pi_E} \right),
\label{eq:obj}
\end{equation}
where $\rho^t_{\pi}(s)$ is the probability that the policy visits state $s$ at time $t$ and $\rho_\pi(s) = \frac{1}{T}\sum_{t=1}^T \rho^t_{\pi}(s)$ is the average visitation across time. In essence, the above objective states that we want the learned policy to visit the same states as the expert. However, optimizing \cref{eq:obj} is challenging as it requires sampling from the learned policy, which can usually only be done accurately by interacting with the environment. 

Instead, the most common approach to imitation learning, Behavior Cloning (BC), reduces the problem to standard supervised learning \citep{ross2010efficient}. Using the \textit{opposite} direction of the KL divergence with respect to \cref{eq:obj}, $\pi_\theta$ can be learned purely offline.
\begin{equation}
\Ls_\text{BC}(\theta) = \E_{s \sim \rho_{\pi_E}}\left[\kl(\pi_E(\cdot | s) || \pi_\theta( \cdot | s)) \right]
\label{eq:bc}
\end{equation}
In this case, we only need samples from $\pi_E$ typically in the form of a dataset of $N$ demonstrations $\D_N = \{\tau_1, \dots, \tau_N\}$, where each demonstration $\tau_i = (s_1, a_1, s_2, a_2, \dots, s_{T_i}, a_{T_i} )$  of length $T_i$ is a valid sequence through the state-action space according to the dynamics. 

Demonstrations are assumed to be sampled from an absolute expert $\pi_E$---however, this assumption in practice is unrealistic. As an example, though we might only care about completing a ``task'' when learning robot policies, there are often several strategies of doing so, and even when using the same strategy, different demonstrators may be subtly different in how they complete the task, which ends up affecting our empirical estimate of the expert. In robot demonstration curation, we ask how we can better define the empirical expert.

\subsection{Demonstration Curation}
While theoretical analyses fix the expert distribution, in practice it is empirically defined by the users and practitioners who collect data. In turn, choices made during data collection can affect the performance of a policy trained with behavior cloning. For example, a novice data collector may produce less predictable actions than an experienced one, and pooling together the data from multiple demonstrators may lead to a more complex action distribution. Moreover, choices made within individual demonstrations $\tau$, such as using differing strategies or varied approaches to complete a task, might make learning from the overall dataset $\D_N$ more difficult. Thus, the problem of demonstration curation in imitation learning is concerned with how we can shape the expert policy distribution $\rho_{\pi_E}$ such that we can attain the highest performance at a given task. Mathematically, we do so by adjusting the empirical expert distribution, $\hat{\rho}_{\pi_E}(s) = \frac{1}{n} \sum_{i=1}^n \frac{1}{T} \sum_{t=1}^T \mathds{1}(s = \tau_{i,t})$ of the dataset, where $\tau_{i,t}$ is the $t$th state of the $i$th demonstration.

We consider the general problem of shaping the empirical expert distribution $\hat{\rho}_{\pi_E}(s)$ tabula rasa at the demonstration level, assuming all demonstrations are successful at completing the desired task. Specifically, our goal is to determine a score function $S(\tau)$ in a purely offline fashion that is able to predict the \textit{quality} of demonstrations, where quality is determined by the performance of a policy trained with behavior cloning on the score-filtered demonstration dataset
\begin{equation}
    \D_N(\kappa, S) = \{\tau_i \mid S(\tau_i) > \kappa,  \forall i = 1, \dots, n \}
    \label{eq:filter_ds}
\end{equation}
for some quality threshold value $\kappa$. This is a more difficult problem than considered in prior work. Data mixing approaches \citet{hejna2024remix} only modify $\hat{\rho}_{\pi_E}(s)$ at the mixture level, i.e. adjusting coarse coefficients $\alpha$ over groups of demonstrations. Instead, considering data curation at the individual demonstration allows us to have a fine-grained understanding of what strategies and expert distributions lead to the best performing policy downstream. Works in interactive data curation necessitate both online access to the environment and expensive oracle feedback  \citep{hoque2021thriftydagger, cui2019uncertainty} for curation. Our setting is purely offline and unsupervised, allowing methods we develop to be applied to virtually any robotics dataset available. However, given we have no explicit signal from the environment in the demonstration curation setting, we aim to define $S$ according to unsupervised objectives, namely mutual information. In the next section, we discuss why mutual information between states and actions can be a valuable scoring function for behavior cloning.


\section{Mutual Information as a Quality Metric}
\label{sec:method:mutual-information}
Mutual information captures the bits of knowledge one gains about one random variable by observing another, in essence measuring predictability. In BC, we want to train a policy $\pi_\theta$ to predict the action $a$ from the state $s$. Thus the mutual information between states and actions is a rather natural choice for a quality metric. In this section we interpret the following factorization of mutual information in the context of robot data curation:
\begin{equation}
    I(S;A) = H(A) - H(A \mid S)
\label{eq:mi}
\end{equation}
where $S$ and $A$ represent random variables for the state and action. First, we will discuss why minimizing the conditional action entropy allows for more accurate policies. Second, we discuss why maximizing the marginal action entropy is required to learn a good policy.

\begin{figure*}
    \centering
    \includegraphics[width=0.925\linewidth]{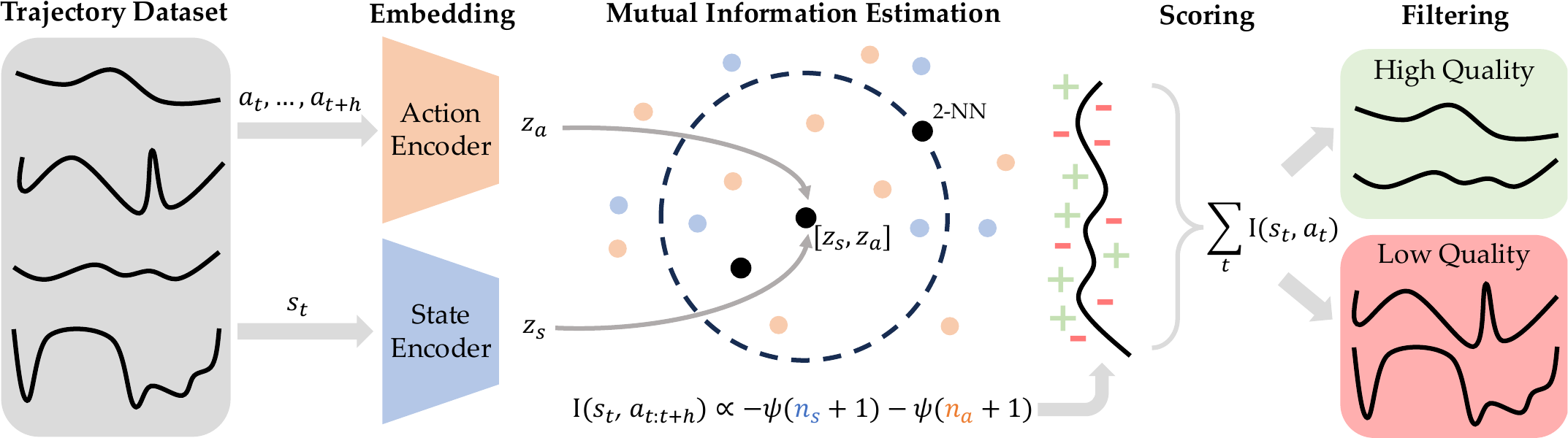}
    \vspace{-0.1in}
    \caption{A graphical depiction of the \abv method. First, we begin by learning VAEs for states and action chunks to produce latent representations $z_a$ and $z_s$. Using these latent representations, we apply the KSG $k$-nearest-neighbor based mutual information estimator. Finally, we filter demonstrations based on their estimated mutual information.}
    \label{fig:method}
    \vspace{-0.2in}
\end{figure*}

\subsection{Minimizing Conditional Action Entropy}
Our overall objective is to align the distribution of the learned policy with that of expert data (\cref{eq:obj}). Following Theorem 4.1 of \citet{belkhale2024data}, we can bound the distribution matching objective from \cref{eq:obj} using the log-sum inequality in terms of the divergence between the learned policy and expert policy at each time step:
\begin{equation*}
    \kl\left(\rho_{\pi_\theta} || \rho_{\pi_E} \right) \leq \frac{1}{T} \sum_{t=1}^T (T - t) \E_{s \sim \rho^t_{\pi_\theta}}\left[\kl\left( \pi_\theta(\cdot | s) || \pi_E( \cdot | s)\right)\right].
\end{equation*}
Intuitively, if we can keep the policies close enough to each other at every state, then we should be able to better reproduce the desired state distribution. Below, we use this fact to argue why low conditional action entropy $H(A \mid S)$ (term 2 in \cref{eq:mi}) leads to better BC performance \citep{belkhale2024data}.

\smallskip \noindent \textbf{Ease of Fit.} Lower entropy distributions are generally simpler, possibly making them easier to match. For example, an action distribution that can only take on a single value has zero conditional entropy. Note that BC (\cref{eq:bc}) optimizes the opposite direction of the KL-divergence with respect to the above abound. The forward and reverse KL-divergences are only equal when $\pi_E$ and $\pi_\theta$ are the same. This is more likely to happen for simple distributions, allowing us to make progress towards the true state matching objective in \cref{eq:obj}. 

\smallskip \noindent \textbf{Multimodality.} Lower entropy distributions often have fewer modes or peaks. Given the forward and reverse KL-divergences have different behaviors around modes, e.g., mode-seeking versus mode-covering, they are more likely to exhibit similar behaviors on unimodal datasets.  

\smallskip \noindent \textbf{Privileged Information.} It can be difficult for a policy to fit demonstrations when the data collector has access to information unavailable to the policy. For example, a data collector may have extra sensory information--such as direct line-of-sight to observe objects that are occluded in the robot's camera views. The resulting actions might only be predictable when given access to the unobserved variable $Z$. Mathematically, we can bound the mutual information between the unobserved $Z$ and actions $A$ by $H(A|S) \geq I(A;Z|S)$ \citep{cover1999elements}. Thus, by minimizing the entropy of the action distribution we ensure that unobserved factors have a smaller effect on the data.

\subsection{Maximizing Marginal Action Entropy}
In addition to minimizing conditional action entropy, mutual information encourages high entropy in the marginal action distribution $H(A)$ (the first term of \cref{eq:mi}). While this might be puzzling at first, it has an important regularizing affect. Without considering the marginal entropy of actions $H(A)$, the conditional entropy $H(A|S)$ could be trivially minimized by distributions that have constant actions, e.g. taking the same action regardless of the state. Having a high marginal action entropy avoids this pitfall, forcing the learned policy to pay attention to the state when making predictions, which is desirable for closed-loop control. 

We find mutual information to be a useful metric for data quality. However, practitioners in robotics have often considered simpler metrics as heuristics for data quality. For example, is often chosen as a convenient proxy for enabling generalization. Mutual information objective $I(S;A)$ can equivalently be decomposed  as $H(S) - H(S|A)$ to uncover state entropy maximization. However, the latter term is less interpretable in this objective. As we will later average mutual information estimates across entire trajectories, \abv will not explicitly optimize for initial state diversity, but instead the underlying predictability of the data.

\section{Method}
\label{sec:method:score}
Though mutual information is perhaps a natural metric for data curation, it can be practically difficult to estimate \citep{beyond-normal-2023}. 
In this section we propose the \fullname (\abv) method for computationally estimating mutual information for demonstration data. Though mutual information is usually considered at the distribution or dataset level, we are interested in scoring individual demonstrations for data curation. Thus, we measure the contribution of individual episodes to the overall mutual information of the dataset. Fortunately, this can easily be done as the majority of of empirical mutual information estimators can be decomposed into an average of sample-wise estimators. 
\begin{equation}
    \hat{I}(S;A) = \frac{1}{|\D_N|} \sum_{(s_i,a_i) \in \D_N} \hat{I}(s_i, a_i ; \D_N)
    \label{eq:empirical_mi}
\end{equation}
As previously outlined in \cref{sec:related_work}, there are several possible neural estimators of mutual information which can be applied to high dimensional robotics data. However, the majority of existing methods like InfoNCE \citep{oord2018representation} and MINE \citep{belghazi2018mine} have extremely high sample requirements for effective estimation which are unrealistic for real world BC datasets. To overcome this challenge we propose \fullname, which uses $k$-nearest-neighbor ($k$-NN) estimates of mutual information. Our method involves three steps -- representation learning, mutual information estimation, and scoring -- which we outline below.

\subsection{Representation Learning} 
As $k$-NN estimators of mutual information do not require training a deep neural network, they have been found to be more sample efficient than other estimators. However, they are typically applied to low-dimensional datasets in contrast with robotics datasets which often contain multiple images and sensors. Directly applying $k$-NN estimators to raw image data may suffer poor performance as distances as become meaningless due to the curse of dimensionality \citep{curseofdim_knn}. To remedy this problem and provide a space suitable for non-parametric estimation we train separate Variational Auto-Encoders (VAEs) \citep{kingma2013auto} to embed both the states and actions into low-dimensional representations. 

We denote embedded states as $z_{s,i} = f_s(s_i)$ and embedded actions as $z_{a,i} = f_a(a_i)$. Though other techniques for representation learning exist, we choose to learn VAEs because they enforce an isotropic Gaussian constraint onto the latent distribution $p(z)$. This is particularly desirable for $k$-NN based mutual information estimators for two reasons. First, enforcing a prior over the latent distribution ensures that distances between embedded states and actions are meaningful -- a necessary prerequisite for statistics based on $k$-NN. Second, $k$-NN based mutual information estimators are commonly assessed on Gaussian distributed data, where they are known to perform well \citep{beyond-normal-2023}. When training the VAEs $f_s$ and $f_a$ we try to select the smallest latent dimension that we believe can sufficiently capture the variable. 

\subsection{k-NN MI Estimation}
Given a latent representation of the states and actions, we can estimate the contribution of an individual state-action pair to the overall mutual information of the dataset using $k$-NN based estimators. The general intuition behind these estimators is that the probability density function around a sample is proportional to how many other data points are near it, which can be measured with the nearest neighbors. If the density function is high near a data point, then we expect there to be many samples around it and thus have a small $k$-NN distance. Conversely, if the density function is low we expect a large $k$-NN distance. Averaging these density estimates allows us to estimate entropy \citep{kozachenko1987sample}, which can be extended to mutual information. In particular, we use the KSG estimator from \citet{kraskov2004estimating}, which we outline below. 

Let $\rho_{k,i}$ be the $k$-NN distance of the $i$th state action pair $[z_s, z_a]$ in the joint space $\mathcal{Z}_\mathcal{S} \times \mathcal{Z}_\mathcal{A}$ defined using the metric
\begin{equation*}
    ||[z_s, z_a] - [z_s', z_a'] || = \max \{||z_s - z_s'||_2, ||z_a - z_a'||_2 \}.
\end{equation*}
The L2 norm between individual latents follows the Gaussian distribution learned by the VAEs. The infinity norm between the $\mathcal{Z}_\mathcal{S}$ and $\mathcal{Z}_\mathcal{A}$ spaces allows the errors from estimates of $S$ and $A$ to cancel in the final KSG estimator. Then, in the context of \cref{eq:empirical_mi}, the KSG estimator is given by:
\begin{equation*}
    \hat{I}(s_i, a_i ; \D_N) \propto -\psi(n(z_{s,i}) + 1) - \psi(n(z_{a,i}) + 1)
\end{equation*}
where $\psi$ is the di-gamma function and $n(z_{s,i)}$ is
\begin{equation*}
    n(z_{s,i}) = \sum_{j \ne i} \mathds{1}\{ ||z_{s,i} - z_{s,j}||_2 \leq \rho_{k,i} \}
\end{equation*}
or the number of latent states $z_s$ less than or equal to the $k$-nearest-neighbor distance $\rho_{k,i}$ in $\mathcal{Z}_\mathcal{S} \times \mathcal{Z}_\mathcal{A}$. The same quantity is analogously defined for actions. We omit constant terms that do not affect the relative contribution of different state-action pairs to the mutual information. We refer the reader to \cref{fig:method} for a pictoral example.

As computing $k$-NN is computationally prohibitive as the dataset size increases, we take a randomized approach. Using a large batch size, we iterate over the dataset multiple times, each time with a distinct shuffling order. We then compute the mutual information contribution $\hat{I}(s, a ; B)$ within each batch $B$ for multiple values of $k$ and average.

\subsection{Scoring}
Given a set of mutual information estimates, our goal is to determine a scoring function $S$ for each episode $\tau$. Intuitively, we can then define the scoring function for each demonstration as the average contribution of that demonstration $\tau$ to the overall mutual information estimator $\hat{I}(S,A)$. 
\begin{equation*}
    S(\tau) = \frac{1}{T} \sum_{t=1}^{T} \hat{I}(s_t, a_t ; \D_N)
\end{equation*}
Since we are filtering datasets by the score, we primarily care about the relative ordering of mutual information estimates rather than their absolute values. In practice, we standardize the dataset by first clipping state-action estimates $\hat{I}(s, a)$ to lie between the 1st and 99th percentiles to prevent excessive influence of outliers.

\begin{figure}
    \centering
    \includegraphics[width=0.68\linewidth]{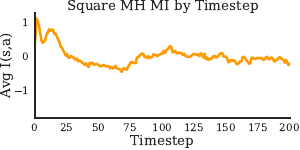}
    \vspace{-0.1in}
    \caption{The average estimated $\hat{I}(s;a)$ per timestep for high quality data (``better'' demonstrators) in ``Square MH'' from RoboMimic \citep{robomimic}. Notice that at the start of the trajectory and after the grasp (75-100 steps), $\hat{I}$ is highest, while it is low during the grasp period (50-75 steps).}
    \label{fig:per_timestep}
    \vspace{-0.2in}
\end{figure}

Note that even though we have scores for each state-action pair $\hat{I}(s, a)$, we do not use them to directly filter the data. Such an approach would not only be noisier, but also remove all parts of a task that are inherently harder to predict, but necessary for success. For example, free motion towards an object is likely easy to predict, but the exact time-step at which the gripper should close is hard to predict. We show this in \cref{fig:per_timestep} for the high quality demonstrations in one dataset, where $\hat{I}(s, a)$ is significantly higher at the start during free motion and lowest when grasping the object ($\sim$ 50--75 steps). Filtering data by mutual information at the state-action level thus might drop data for crucial parts of a task that inherently have lower mutual information in favor of easily predictable motion.

Using the score function $S$, we can subset the dataset to include only demonstrations that contribute positively towards the average mutual information estimate of the dataset.

\section{Experiments}
\begin{figure}
    \centering
    \includegraphics[width=\linewidth]{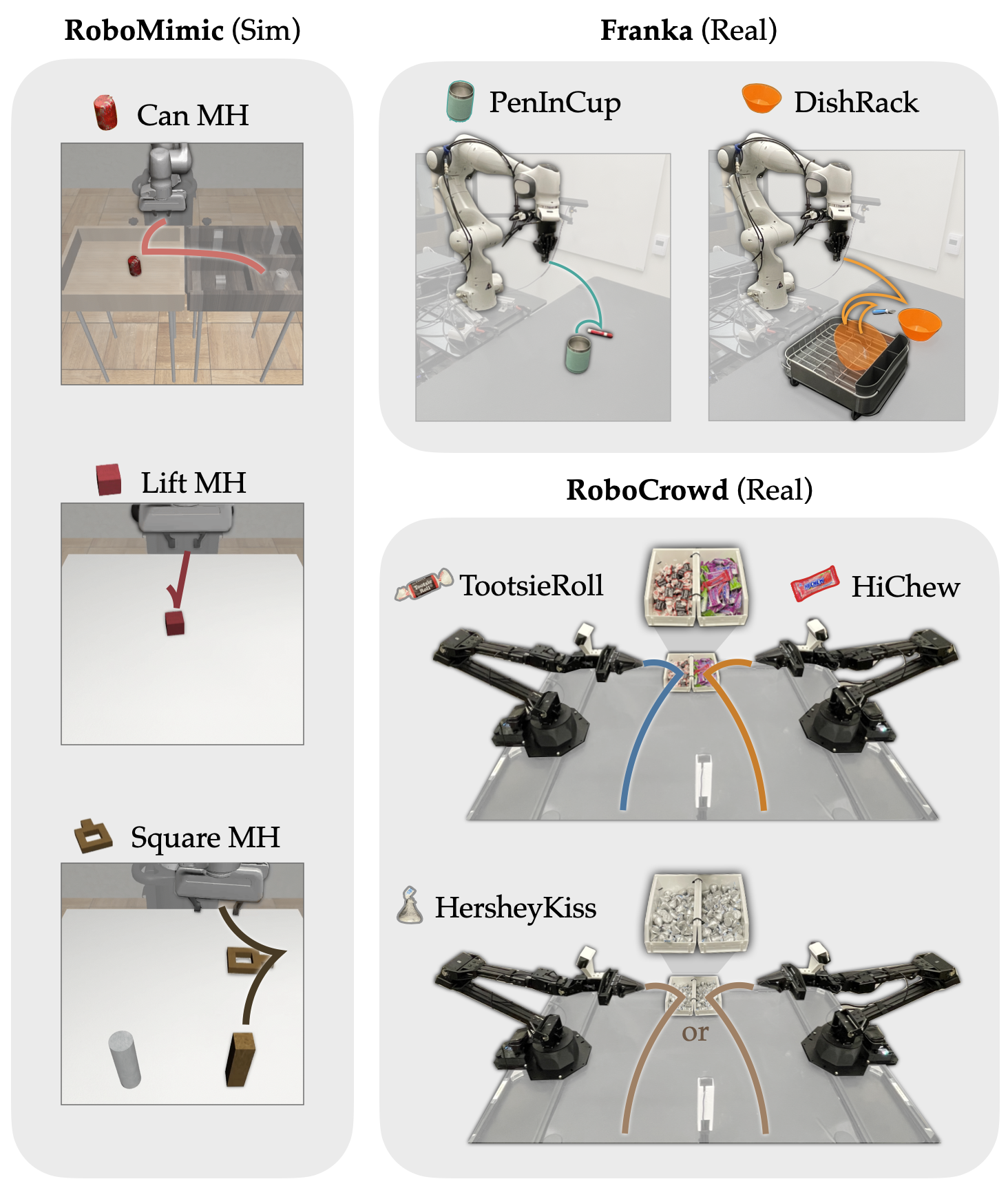}
    \vspace{-0.2in}
    \caption{Visualization of the tasks represented in the datasets we use in this work, including the Can MH, Lift MH, and Square MH datasets from RoboMimic; real-world PenInCup and DishRack datasets collected on a Franka robot; and the real-world TootsieRoll, HiChew, and HersheyKiss datasets from RoboCrowd for the ALOHA robot.}
    \label{fig:datasets}
\end{figure}

We aim to answer the following questions: (1) How well does \abv curate robot data? (2) How do different mutual information estimators affect performance? (3) Can data curation via mutual information improve performance on downstream policy learning? and (4) What is important to \abv's performance? Additional results are presented in \cref{app:results}.

\begin{figure*}[t]
    \centering
    \includegraphics[width=0.49\linewidth]{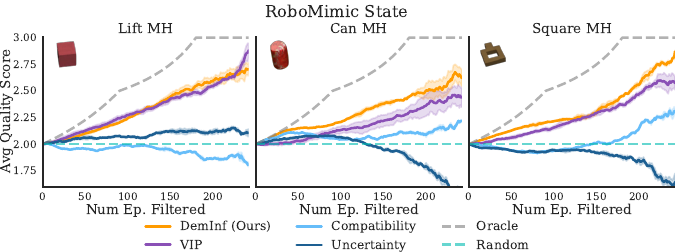}
    \includegraphics[width=0.49\linewidth]{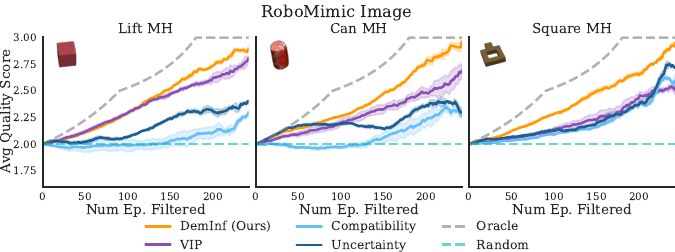}
    \vspace{-0.15in}
    \caption{Average quality of demonstrations remaining in datasets after filtering with different choices of $S$ on the Lift, Can, and Square Multi-Human (Mh) datasets from the Robomimic benchmark with states (Left) and images (right). Results are shown as an average of 3 seeds.}
    \label{fig:robomimic_baseline}
\end{figure*}

\begin{figure*}[t]
    \centering
    \includegraphics[width=\linewidth]{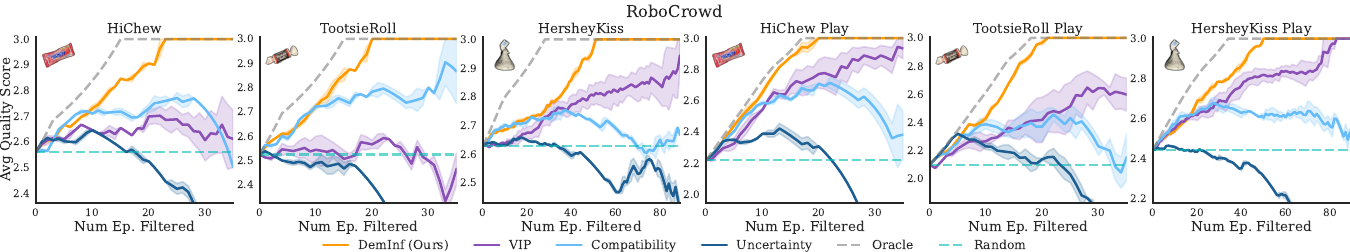}
    \caption{Average quality of demonstrations remaining in datasets after filtering with different choices of $S$ on the Hi-Chew, Tootsie-Roll, and Hershey-Kiss crowdsourced datasets from the RoboCrowd benchmark. We include results for datasets with a combination of expert and only task-relevant data (left), and a version of the data that contains additional unstructured play data (right).  Results are shown as an average of 3 seeds.}
    \label{fig:robocrowd_image}
\end{figure*}

\subsection{Experimental Setup}

\subsubsection{Datasets}
To assess the performance of different robot demonstration curation techniques, we perform experiments on a broad set of datasets spanning simulated, real single-arm, and real bi-arm robots with varying levels of data quality as depicted in \cref{fig:datasets}. Notably, we use datasets where human experts have provided quality labels, allowing us to easily assess different demonstration curation metrics:

\noindent \textbf{RoboMimic.} The multi-human datasets from the RoboMimic benchmark \citep{robomimic} include 100 demonstrations from each of three robot operators for three tasks in increasing difficulty: ``Lift'' where the robot simply lifts a cube, ``Can'' where the robot moves a can from one bin to another, and ``Square'' where the robot places a nut onto a peg. RoboMimic provides quality labels for each operator, which we use to assign quality scores (with scores of 1, 2, and 3 for the ``worse'', ``okay'', and ``better'' demonstrations respectively). We measure the performance of different data curation methods from both state, in which ground truth object information is provided, as well as third-person images. 

\noindent \textbf{Franka.} Using the setup from \citet{droid} with a Franka Panda robot we collect 60 and 80 demonstrations for each of two tasks, ``PenInCup'' and ``DishRack'' respectively. Within each task, we collect 50\% expert demonstrations (quality 1) and 50\%  poor demonstrations (quality 0), where the operator intentionally makes a mistake (e.g. dropping an object, taking a long inefficient path, jerky motion). We use a single third person camera and a wrist camera to train policies and action chunks of size 4. 

\noindent \textbf{RoboCrowd.} The RoboCrowd benchmark from \citet{mirchandani2024robocrowd} contains crowdsourced robot data on the bimanual ALOHA \citep{zhao2023aloha} platform from real, novice users in a natural environment. Data in RoboCrowd varies widely in quality -- many trajectories contain suboptimal data or sequences of ``play'' data that are irrelevant to the target task. RoboCrowd serves as a suitable platform to study data curation as it has a small number of expert demonstrations for each task and human expert quality labels ranging from 0 to 3 for all crowdsourced data. Specifically, we use the ``HiChew Play,'' ``TootsieRoll Play,'' and ``HersheyKiss Play'' datasets which contain both expert demonstrations and crowdsourced demonstrations for candy bin-picking tasks. Every demonstration contains some amount of task-relevant data, with the potential of irrelevant play data in the crowdsourced demonstrations as well. We additionally evaluate on versions of these datasets (``HiChew'', ``TootsieRoll', ``HersheyKiss'') where the unstructured play data has been removed, but where demonstrations still contain task-relevant data of varying quality. The HiChew and TootsieRoll datasets contain 40 demonstrations each and the HersheyKiss dataset contains 100 demonstrations, half of which are expert demonstrations. We use the wrist cameras, overhead camera, and action chunks of size 10 for data curation. \looseness=-1

\subsubsection{Baselines}
We compare against a number of different data quality estimators from prior work in addition to a number of alternative mutual information estimators, which we label with ``(MI)''.

\noindent \textbf{Uncertainty.} Following prior works in active learning for imitation learning \citep{cui2019uncertainty, hoque2021thriftydagger}, we select data based on the uncertainty of an ensemble of 5 policies. Note that while this metric makes sense for active learning, it does not necessarily make sense in the offline setting, and in some ways may be inversely correlated with quality if the ensemble converges better on high quality data. 

\noindent \textbf{Compatibility.} Following \citet{ghandi2023eliciting}, we use a measure of demonstration ``compatibility'' to score data. Namely, a demonstration is compatible with respect to a policy if it has either high ``novelty'' as measured by the prediction variance of an ensemble, or low novelty and high likelihood as measured by the average loss. Though this method was originally designed to be used in the online setting, we adopt it to the offline setting by training a policy on all data, then estimating the ``compatibility'' for each demonstration with respect to the overall policy.  

\noindent \textbf{VIP.} Value Implicit Pre-training \citep{ma2023vip} is an action-free method that leverages the dual formulation of the goal-conditioned RL problem to learn a ``universal'' value function. We use VIP to estimate data quality by considering the total predicted reward over a demonstration. 

\noindent \textbf{InfoNCE (MI).} We use the symmetric InfoNCE \citep{oord2018representation} objective used to train CLIP \citep{radford2021learning} which converges to an estimate of mutual information. We compare to InfoNCE as CLIP is commonly used to curate datasets in vision and language \citep{schuhmann2022laion}.

\noindent \textbf{MINE (MI).} MINE \citep{belghazi2018mine} leverages the dual form of the KL divergence to estimate the mutual information using a learned critic function.

\subsubsection{Architectures}
For all state-based experiments we use MLPs with two hidden layers of size 512. For image-based experiments we use ResNet-18 Encoders \citep{he2016deep} with spatial softmax activations following \citet{robomimic}, which are concatenated with state information as input to a MLP with two hidden layers of size 1024. When training VAEs from images we use matching ResNet-18 Decoder networks for each view. For each dataset we use the same architecture for all methods, where the latent $z$ dimension is set to be consistent across both \abv and baselines. Images resized to $84 \times 84$ for RoboMimic and $128 \times 128$ otherwise. For all experiments we use the Adam optimizer with learning rate 0.0001 and a batch size of 256. State-based models are trained for 50,000 steps and image based models are trained for 100,000 steps using VMs provided by a Google TPU Research Cloud Grant. We run three seeds for all methods. More details and hyper-parameters can be found in \cref{app:implementation}.

\begin{figure}
    \centering
    \includegraphics[width=0.7\linewidth]{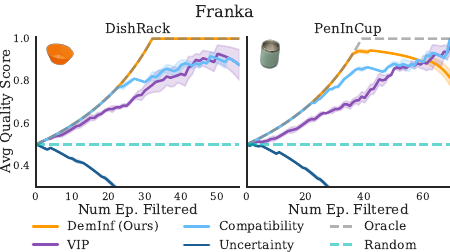}
    \vspace{-0.1in}
    \caption{Average quality of demonstrations remaining in datasets after filtering with different choices of $S$ on the Franka Datasets. Average of 3 seeds.}
    \label{fig:franka_baseline}
\end{figure}

\begin{figure*}[t]
    \centering
    \includegraphics[width=0.495\linewidth]{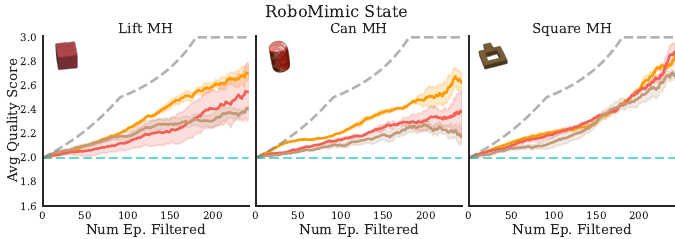}
    \includegraphics[width=0.495\linewidth]{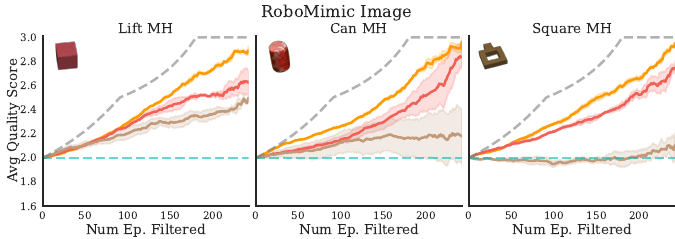}
    \includegraphics[width=\linewidth]{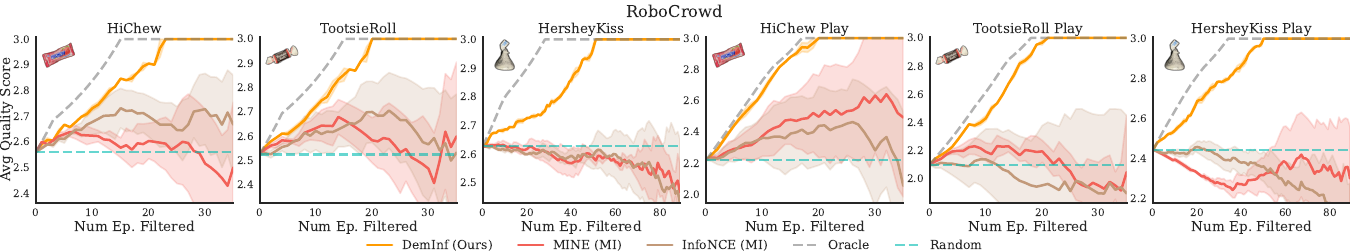}
    \vspace{-0.225in}
    \caption{Average quality of demonstrations remaining in datasets across RoboMimic and RoboCrowd after filtering with different mutual information estimators. Again, all experiments are averaged over 3 seeds. We found InfoNCE and MINE to exhibit higher variance than \abv and struggle with higher dimensional inputs, especially with lower amounts of data.}
    \label{fig:mi_estimators}
\end{figure*}

\subsection{How well does \abv curate data?}

To assess how well \abv can curate data, we plot the number of episodes filtered from each dataset against the average resulting expert quality label. This amounts to considering every possible dataset generated by sweeping the threshold level $\kappa$ in the sub-setted dataset according to scoring function $S$ (see \cref{eq:filter_ds}). Doing so allows us to simultaneously assess how well each method would does at every threshold level. The closer the curve is to the ``oracle'' or curating directly by the expert labels, the better. Note that one should not over-index on the right-hand size of the plot as with a typical learning curve, as that represents performance only as the dataset reaches 10\% of its original size.

\noindent \textbf{State-Based Results.}
We depict results on the state-based RoboMimic benchmark on the left side of \cref{fig:robomimic_baseline}. \abv performs as well or better than baselines in all environments, though there is not a particularly large gap with VIP. 

\noindent \textbf{Image-Based Results.}
In the image-based settings we find that \abv performs even better, surpassing all methods on both RoboMimic (\cref{fig:robomimic_baseline}) and Franka (\cref{fig:franka_baseline}). On the ``DishRack'' task, \abv is able to exactly match the oracle. VIP performs comparably worse in this setting, likely because its bootstrapping-based RL objective is more difficult to optimize in higher dimensions. Conversely, the uncertainty based metrics perform better in RoboMimic. The  compatibility metric performs quite well on the Franka tasks, likely because the low quality data was explicitly collected with higher entropy, making it easy to distinguish with policy loss alone.

\noindent \textbf{Crowdsourced Data.}
To assess \abv's ability to filter data from a wide variety of operators, styles, and quality levels we turn to the RoboCrowd benchmark. We again find that \abv most consistently filters out low-quality data with respect to the expert labels. The extreme diversity of these datasets, combined with the limited number of demonstrations available (40-100) proves extremely challenging for all baselines, which often provide only a small edge over random sampling. Selecting based on uncertainty performs quite poorly here -- demonstrating that when learning offline, uncertainty is a poor metric, and certainty (its inverse) may perform better. These results suggest that methods designed for active learning and interactive data collection are not sufficient for the problem of offline data curation. When comparing the left side of \cref{fig:robocrowd_image} with the right side, we see that VIP is able to perform better on the ``Play'' datasets can contain task-irrelevant sequences in the demonstrations. While this might be counter-intuitive at first, VIP is goal-conditioned and scores state, next-state tuples based on perceived progress towards the goal. Thus, data with large amounts of irrelevant data extending the length of trajectories, VIP has an easier time filtering.

\subsection{Mutual Information Estimators}

\cref{fig:mi_estimators} shows the performance of different mutual information estimators across RoboMimic and RoboCrowd. While InfoNCE and MINE perform acceptably in state-based settings, they begin to perform significantly worse in image based settings as the dimensionality of the data increases. InfoNCE in particular performs far worse in RoboMimic, underscoring the raw amount of data needed to train a high quality contrastive representation as documented by prior work. Both MINE and InfoNCE perform poorly in the more data-limited regime in RoboCrowd while \abv, which uses non-parametric estimation no top of representation learning, is able to retain performance. Moreover, we find that \abv exhibits far lower variance across seeds, while the parametric estimators were more unstable and had one or two runs that performed far worse than the others. This is particularly problematic for downstream data curation, as one often does not have ground truth labels to check the quality of the scoring function. \looseness=-1

\begin{figure}
    \centering
    \includegraphics[width=\linewidth]{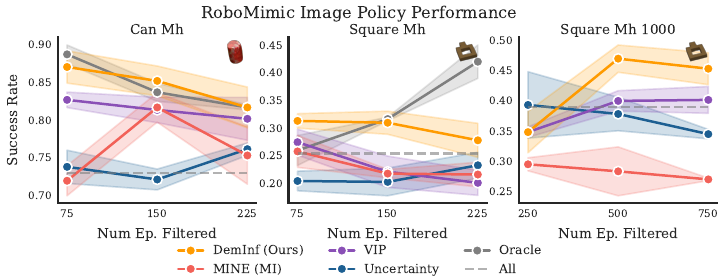}
    \vspace{-0.25in}
    \caption{Performance of ResNet18 + MLP BC Policies trained on filtered subsets of RoboMimic from Images. Evaluations are averaged over 200 trials for each of 3 seeds (600 total) after 100K training steps. Each dataset begins with 300 demonstrations, except for ``Square Mh 1000'' which has 1000.}
    \label{fig:robomimic_policy}
    \vspace{-0.2in}
\end{figure}

\begin{figure*}[t]
    \centering
    \includegraphics[width=0.95\linewidth]{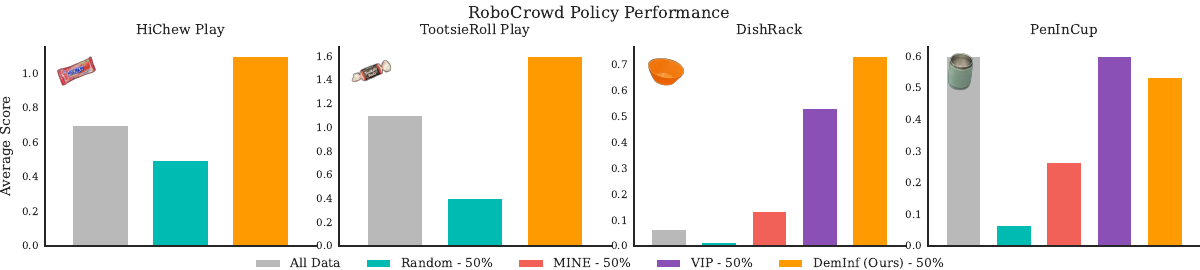}
    \vspace{-0.1in}
    \caption{Performance of ACT trained on filtered versus non-filtered RoboCrowd (HiChew and TootsieRoll) and Diffusion Policy trained on filtered versus non-filtered data for the Franka tasks (DishRack and PenInCup). Evaluations for RoboCrowd are shown using the same scoring procedure from \citet{mirchandani2024robocrowd} over 10 trials, and evaluations for Franka are computed using binary success over 15 trials.}
    \label{fig:robocrowd_policy}
\end{figure*}

\subsection{Does demonstration curation affect policy performance?}
While comparing to ground truth labels allows us to assess the quality of different approaches to filtering, we ultimately care about the performance of downstream BC policies. In \cref{fig:robomimic_policy} we train BC policies on RoboMimic Can MH and Square MH from images when filtering different numbers of episodes from the dataset $\D_N$ according to the best baselines for the tasks. Overall, we find that at all data scales \abv performs better than baselines, which exhibit far less consistent performance trends overall. For example, uncertainty often shows little improvement until the majority of the dataset is filtered. Crucially, we see that filtering data with \abv performs \emph{better} than training on all of the data by over 10\% in Can and is the only method to improve upon training on all of the data in Square. To further assess performance on larger datasets, we add an additional 700 successful rollouts from Diffusion Policy~\citep{chi2023diffusion} to Square MH to bring the total to 1000 demos for ``Square MH 1000''. Even with larger datasets, \abv maintains performance. We include results for robomimic state in \cref{app:results}.

This trend continues in real world evals for both RoboCrowd and Franka where we compare training policies on all of the demonstrations (no filter), training policies on a random 50\% subset of the demonstrations, and filtering 50\% of the demonstrations with \abv in \cref{fig:robocrowd_policy}. We train ACT \citep{zhao2023aloha} for RoboCrowd and Diffusion Policy \citep{chi2023diffusion} for Franka. In RoboCrowd, after scoring trials according to the methodology in \citet{mirchandani2024robocrowd} (1 for grasping any number of candies, 2 for returning it to the user, and 3 for returning only one as in the demonstrations), we find that the \abv policy not only more commonly successfully completes the task, but also exhibits better motion when compared to training on all of the data. When considering the same number of demonstrations randomly selected from the dataset, we find that the gap in score is larger, indicating that better performance can be attained by collecting only good data. On the Franka tasks \abv always outperforms the random subset, but training on all data does slightly better than filtering with \abv on ``PenInCup''. This is likely because ``PenInCup'', being the harder task, requires more data for representation learning even if the quality is poor. However, when dataset size is kept at parity (50\% random subset) we see a huge difference in performance, indicating that in all tasks data quality matters. VIP does well on the Franka tasks as well, which can be attributed to the fact that the purposefully suboptimal demonstrations we collected were significantly longer than the expert demonstrations, making filtering via a value function particularly effect.

\begin{figure}
    \centering
    \includegraphics[width=0.75\linewidth]{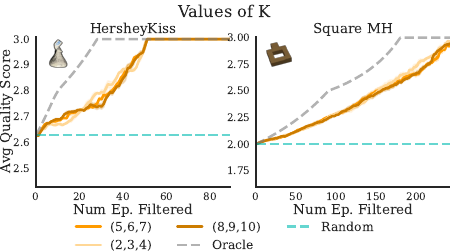}
    \vspace{-0.1in}
    \caption{Performance of 
    \abv with different ranges of $k$ for $k$-NN}
    \label{fig:ks_ablation}
    \vspace{-0.1in}
\end{figure}

\begin{figure}
    \centering
    \includegraphics[width=\linewidth]{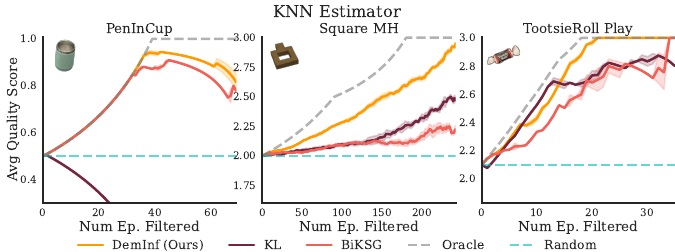}
    \vspace{-0.15in}
    \caption{Performance of 
    \abv with different $k$-NN mutual information estimators}
    \label{fig:knn_ablation}
    \vspace{-0.1in}
\end{figure}

\subsection{Ablations}
Finally, we consider which design choices of \abv affect its ability to curate demonstration data. Here, we consider the value of $k$ used for $k$-NN and the type of non-parameteric mutual information estimator. In \cref{app:results} we include additional ablations over the size of the latent dimensions of $z_s$ and $z_a$ and the value of $\beta$ for the VAEs. \cref{fig:ks_ablation} shows the performance of \abv with different ranges of $k$ used to compute the mutual information. We average final predictions over this range. \abv's performance is generally robust to this parameter, with no substantial change in performance in both HersheyKiss and Square MH. However, the story is different for the choice of $k$-NN estimator. As shown in \cref{fig:knn_ablation}, we found the KSG estimator from \citet{kraskov2004estimating} to be superior to both the BiKSG estimator from \citet{belghazi2018mutual} and the na\"ive application of the differential entropy estimator from \citet{kozachenko1987sample} (KL). This indicates that the quality of the latent space, as well as the quality of the estimator, are important for downstream performance. \looseness=-1
\section{Conclusion}
\label{sec:conclusion}

In this work, we propose the \fullname (\abv) procedure as a method for data curation in robot imitation learning. Specifically, we motivate mutual information as a useful basis for measuring the quality of individual demonstrations, and instantiate mutual information estimators as a way to rank and select demonstrations. Across several datasets of human-teleoperated demonstrations in both the real-world and simulation, we find that the \abv outperforms several prior methods at measuring the quality of demonstrations. \looseness=-1

\subsection{Limitations}

\noindent \textbf{Time Evolution.} \abv considers the aggregate mutual information between states and actions within the dataset. Our analyses, however, largely ignores the fact that these quantities are linked in the sequential setting via the environment dynamics. Concretely, we can increase the marginal action entropy with a more random policy, which will likely increase the state entropy as well through the dynamics. However, this comes at the cost of increasing $H(A|S)$, and it is unclear how \abv balances these factors. An approach to disentangle the effects of dynamics might consider measuring $I(S_1;A_1, \dots, A_T)$, or the mutual information between initial states and action sequences. However, doing so reduces the amount of data available for estimation by a factor of $T$, which is typically 100-1000, making this approach more challenging in practice.

\noindent \textbf{Pauses.} Because \abv considers the average estimated $\hat{I}$ across a trajectory, it is susceptible to preferring data that is predictable, but might not make progress towards completing the task. For example, if a robot pauses for an extended amount of time at a particular state, the action distribution is very predictable. However, this behavior is not desired in practice. To mitigate this effect, we recommend ensuring that all data completes the task and pauses are filtered.

\noindent \textbf{Greediness.} Note that \abv's curation procedure is greedy and not globally optimal -- once we remove an episode we have changed the data distribution, which in turn affects the true mutual information. However, re-running the mutual information estimator on the entire dataset for each filtered demonstration would be far more computationally expensive.

\noindent \textbf{Success.} \abv assumes that all provided demonstrations are successful at completing the desired task.

\subsection{Future Work}
Exciting avenues for future work remain. For instance, extending \abv to the multi-task setting will require disentangling task conditioning from mutual information estimation as to not retain only the easiest tasks. This problem becomes harder in settings where task definitions are not enumerable, like natural language. Other directions include scaling \abv to larger datasets such as \citet{openx} and \citet{droid} to curate better subsets for the robot learning community. Finally, integrating \abv into an online data collection interface could improve data collection efficiency. Though there is more work to do, we believe \abv is a step towards addressing the data problem in robotics. \looseness=-1

\section*{Acknowledgments}
Compute for this research was provided by a Google TPU Research Cloud Grant. This work was supported by ONR project N00014-22-1-2293. We would also like to thank the Stanford IRIS Lab for sharing their ALOHA robot for our final real world experiments and Kaylee Burns for providing feedback and help with set up.


\bibliographystyle{plainnat}
\bibliography{references}

\newpage
\onecolumn
\appendix

\subsection{Extended Results}
\label{app:results}
Here we provide results and ablations that could not fit in the main text.

\noindent \textbf{Additional Results}

\begin{figure}[H]
    \centering
    \includegraphics[width=0.4\linewidth]{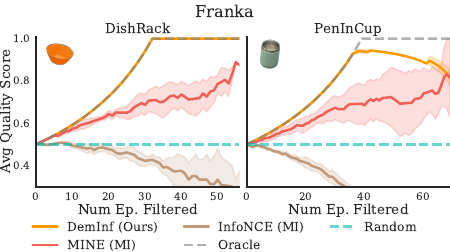}
    \vspace{-0.1in}
    \caption{The performance of different mutual information estimators on the Franka Datasets, cut from the main text due to space.}
\end{figure}

\begin{figure}[H]
    \centering
    \includegraphics[width=0.5\linewidth]{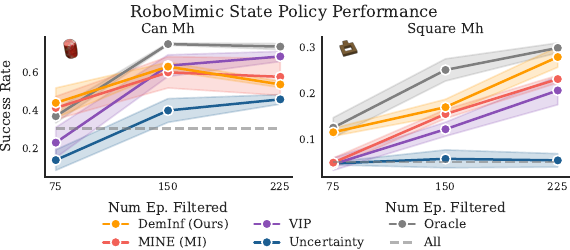}
    \vspace{-0.1in}
    \caption{RoboMimic Policy learning performance from state.}
\end{figure}

\noindent \textbf{Additional Ablations}

\begin{figure}[H]
    \centering
    \includegraphics[width=\linewidth]{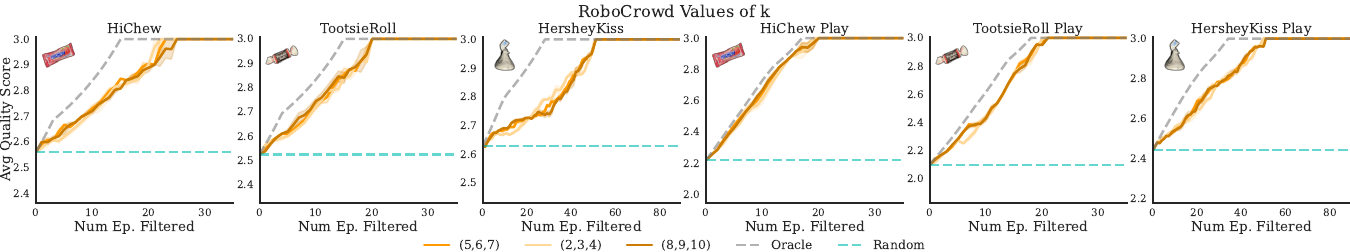}
    \vspace{-0.1in}
    \caption{The effect of different values of $k$ on RoboCrowd}
\end{figure}

\begin{figure}[H]
    \centering
    \includegraphics[width=0.6\linewidth]{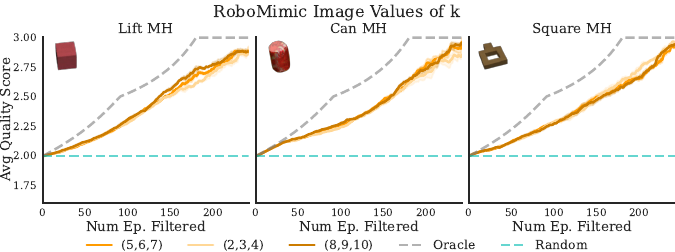}
    \vspace{-0.1in}
    \caption{The effect of different values of $k$ on RoboMimic Image}
\end{figure}

\begin{figure}[H]
    \centering
    \includegraphics[width=0.6\linewidth]{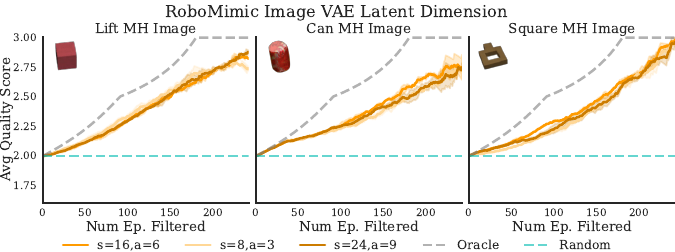}
    \vspace{-0.1in}
    \caption{The effect of different latent dimension sizes for $z_s$ and $z_a$ on RoboMimic Image. we find that performance is relatively robust to this parameter. Unlike all others, this experiment was run over 2, not 3, seeds.}
\end{figure}

\begin{figure}[H]
    \centering
    \includegraphics[width=0.6\linewidth]{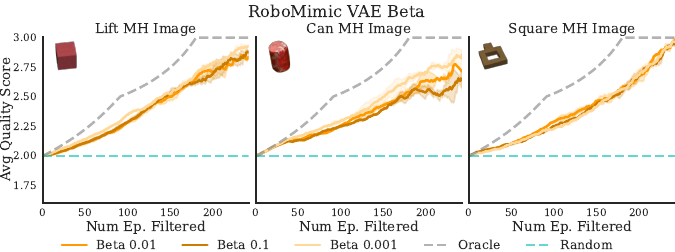}
    \vspace{-0.1in}
    \caption{The effect of different values of VAE $\beta$ on RoboMimic Image. We find that performance is relatively robust to this parameter for RoboMimic. Unlike all others, this experiment was run over 2, not 3, seeds.}
\end{figure}

\begin{figure}[H]
    \centering
    \includegraphics[width=\linewidth]{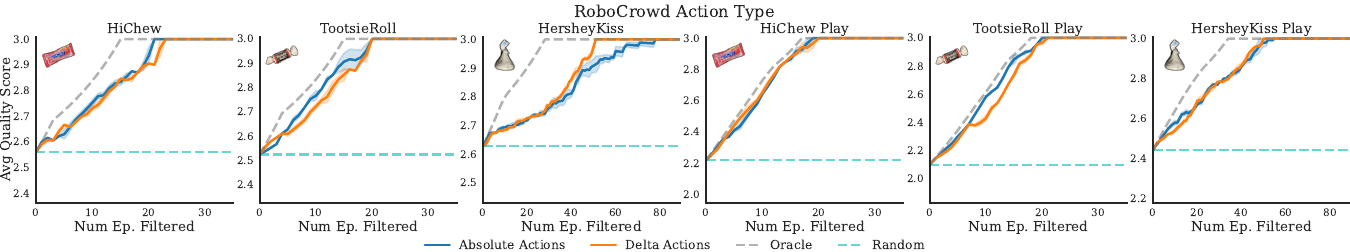}
    \vspace{-0.1in}
    \caption{The effect of using different action spaces for the RoboCrowd dataset.}
\end{figure}

\noindent \textbf{Plots with All Baselines and Estimators.} The below plots show the performance of all methods on the same exact plot, allowing for direct comparison. We additionally consider another baseline ``Policy Loss'', which simply measures the loss of a BC policy.

\begin{figure}[H]
    \centering
    \includegraphics[width=0.495\linewidth]{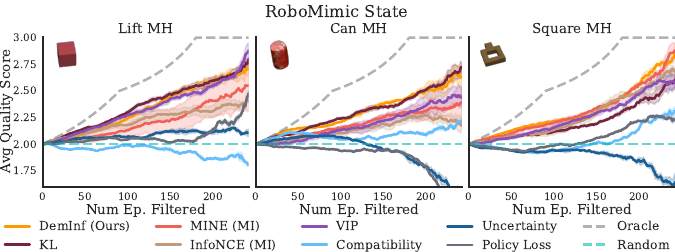}
    \includegraphics[width=0.495\linewidth]{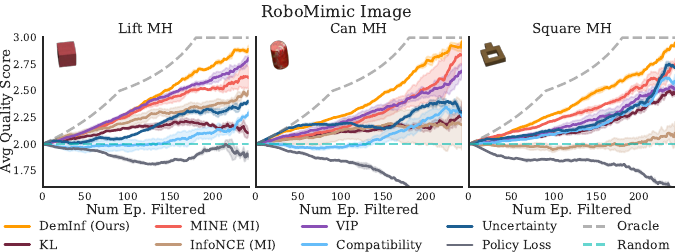}
    \vspace{-0.1in}
    \caption{RoboMimic results for all methods.}
\end{figure}

\begin{figure}[H]
    \centering
    \includegraphics[width=\linewidth]{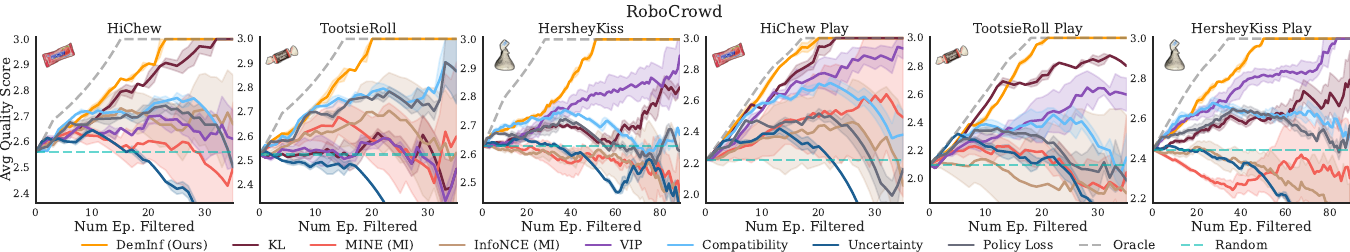}
    \vspace{-0.1in}
    \caption{RoboCrowd results for all methods.}
\end{figure}

\begin{figure}[H]
    \centering
    \includegraphics[width=0.5\linewidth]{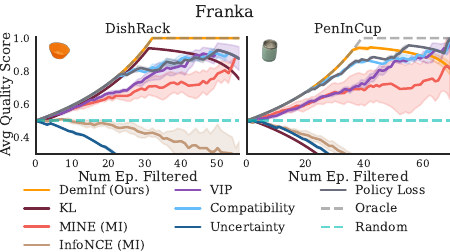}
    \vspace{-0.1in}
    \caption{Franka results for all methods.}
\end{figure}

\subsection{Method Details}
Here we provide more details on each of the different methods used to for data curation. For each method we train the requisite model(s), then run inference over the whole dataset. As we filter at the demonstration level, we aggregate scores for all methods over each demonstration. Mathematically, all scoring functions take the form:
\begin{equation*}
    S(\tau) = \frac{1}{T} \sum_{t=1}^{T} h(s_t, a_t ; \D_n)
\end{equation*}
where $h$ is some function of the state-action pairs in a set of data $\D$ comprised of trajectories $\tau$. We use the subscript $t$ to denote that we index over the steps of a trajectory $\tau$. In practice we clip state-action scores from $h$ at the 1st and 99th percentile. Below we provide the scoring function used for all methods.

\vspace{0.075in}

\noindent \textbf{\fullname (\abv)}. We first fit an action VAE $z_a = f_a(a)$ and state VAE $z_s = f_s(s)$. Then, we iterate over the entire dataset 4 times, computing scores in random batches of size 1024. The score function is then
\begin{equation*}
        h(s_i, a_i; B) = \hat{I}(s_i, a_i ; B) \propto -\psi(n(z_{s,i}) + 1) - \psi(n(z_{a,i}) + 1)
\end{equation*}
where $B$ is a random batch and $n$ is defined as \cref{sec:method:score}.

\vspace{0.075in}
\noindent \textbf{BiKSG}. We follow the same approach as in \abv, except the mutual information is estimated as
\begin{equation*}
        h(s_i, a_i; B) = \hat{I}(s_i, a_i ; B) \propto -\log n(z_{s,i}) - \log n(z_{a,i})
\end{equation*}
and we use the $l2$ distance metric over $\mathcal{Z}_\mathcal{S} \times \mathcal{Z}_\mathcal{A}$ without the $l_\infty$ norm:
\begin{equation*}
    ||[z_s, z_a] - [z_s', z_a'] || = ||[z_s, z_a] - [z_s', z_a'] ||_2
\end{equation*}

\vspace{0.075in}
\noindent \textbf{KL}. We follow the same approach as in \abv, except the mutual information is estimated using separate terms for $H(S)$, $H(A)$ and $H(S, A)$ where each term is given by the differentiable entropy estimator from \citet{kozachenko1987sample}. Let $z^k_{s,i}$ be $z_{s,i}$'s $k$-nearest-neighbor. Then, the estimator is given as
\begin{equation*}
        h(s_i, a_i; B) = \hat{I}(s_i, a_i ; B) \propto \log ||z_{s,i} - z^k_{s,i}||_2^{|\mathcal{Z}_\mathcal{S}|} + \log ||z_{a,i} - z^k_{a,i}||_2^{|\mathcal{Z}_\mathcal{A}|} - \log ||[z_s, z_a]_{i} - [z_s, z_a]^k_{i}||_2^{|\mathcal{Z}_\mathcal{S}| + |\mathcal{Z}_\mathcal{A}|}
\end{equation*}
where $|\mathcal{Z}_\mathcal{S}|$ is the dimension of the latent space.

\vspace{0.075in}
\noindent \textbf{MINE}. MINE optimizes a critic function $f_\theta(s,a)$ to predict the mutual information using the objective
\begin{equation*}
    \max_\theta \E_{(s,a)\sim \D}[f_\theta(s,a)] - \log \left(\E_{s \sim \D, a \sim D}[e^{f_\theta(s,a)}\right)
\end{equation*}
where the first term is sampled from the joint and the second is sampled from the marginals. The scoring function is then simply:
\begin{equation*}
    h(s_i, a_i; B) = f_\theta(s_i, a_i)
\end{equation*}
In practice MINE uses an exponential moving average of gradient's denominator to un-bias the estimator. We refer to this parameter as $\alpha$ as in the original paper and leave it at 0.9.

\vspace{0.075in}
\noindent \textbf{InfoNCE}. We optimize the symmetric InfoNCE objective from CLIP, which converges to the mutual information up to a constant \citep{ma2018noise}. To do so, we train a state encoder $f_s$ and an action encoder $f_a$. After training, the scoring function becomes:
\begin{equation*}
    h(s_i, a_i; B) = f_s(s_i) \cdot f_a(a_i)
\end{equation*}
or simply the dot product between the two representations.

\vspace{0.075in}

\noindent \textbf{VIP}. VIP \citep{ma2023vip} uses the dual form of the goal-conditioned RL problem, with the negative L2 distance between encoded states as a proxy for the value function $V(s,g) = - ||f(s) - f(g)||_2$. The VIP training objective is 
\begin{equation*}
    \min_\theta \E_{s_1 \sim \rho_1, g \sim D}[||f_\theta(s_1) - f_\theta(g)||_2] + \log \E_{s_t, s_{t+1}, g \sim \D}[\exp\left(||f_\theta(s_t) - f_\theta(g)||_2 - \mathds{1}\{s_t = g\} - \gamma ||f_\theta(s_{t+1}) - f_\theta(g)||_2 \right)]
\end{equation*}
Then, using the learned value function we estimate the ``reward'' of each transition by
\begin{equation*}
     h(s_t, s_{t+1}, g; B) = -||f_\theta(s_{t+1}) - f_\theta(g)||_2 + ||f_\theta(s_{t}) - f_\theta(g)||_2
\end{equation*}
which captures the progress of the transition towards the goal. During training we sample goals uniformly from the future, but during quality estimation we set the goals to be the final state in each demonstration.

\vspace{0.075in}
\noindent \textbf{Compatibility.} Following \citet{ghandi2023eliciting} we train an ensemble of 5 policies. Then, the compatibility score is estimated as:
\begin{equation*}
    h(s_i, a_i; B) = \begin{cases}
        1 - \min \left(\text{L2Loss}(\pi_\theta(s_i), a_i) / \lambda, 1 \right) & \text{if} \text{ std}(\pi_\theta(s_i)) < \eta \\
        1 & \text{otherwise}
    \end{cases}
\end{equation*}
where L2Loss is the average L2 loss of the ensemble and std is the standard deviation of the predictions.

\vspace{0.075in}
\noindent \textbf{Uncertainty.} The uncertainty score is estimated from the same ensemble of 5 policies by the standard deviation of the predictions:

\begin{equation*}
    h(s_i, a_i; B) = \text{std}(\pi_\theta(s_i))
\end{equation*}

\vspace{0.075in}
\noindent \textbf{Policy Loss.} The Policy Loss metric is simply the negative L2 Loss of the network, such that demonstrations with lower loss have a higher score.

\begin{equation*}
    h(s_i, a_i; B) = -\text{L2Loss}(\pi_\theta(s_i), a_i)
\end{equation*}

\subsection{Implementation Details}
\label{app:implementation}
\noindent \textbf{Architectures.} We use the same architectures for all methods whenever possible. For state-based experiments we simply use MLPs with two hidden layers of size 512 with ReLU activations. When training BC policies, we add dropout of 0.5 as we found it to be important to performance. For VAEs we use a symmetric decoder.

For Image experiments, we use ResNet18 architectures followed by a spatial softmax layer, similar to the original setup in \citet{robomimic}. We concatenate representations from all cameras along with the state information, and then feed that to information to an MLP. For RoboMimic we use a three layer MLP with hidden dimension of size 512. For Franka and RoboCrowd we use an MLP with two hidden layers of size 1024. For all methods using a state encoder, we use this architecture. For BC policies we ensemble the MLP, add dropout and use the L2 Loss function for training. MINE additionally concatenates the action before the MLP and InfoNCE trains a separate action encoder using just the MLP architecture. For action encoders and decoders, we use the same architecture as for state. For training VAEs on images, we use the same architecture but in reverse, with ResNet18 Decoders. 

We trained all models on TPU v4-8 VMs provided by the Google TPU Research Cloud. Training for 100K steps took approximately 30 minutes for VAEs and 1 hour for other methods.

\vspace{0.05in}
\noindent \textbf{Hyperparameters}
We set hyper-parameters consistently across settings, e.g. RoboCrowd and try to choose the same parameters for all methods when possible. Hyperparameters for all methods are shown in \cref{tab:hyperparameters}.

\begin{table*}[ht]
\centering
{\renewcommand{\arraystretch}{1.05} 
\begin{tabular}{cccccc}
\textbf{Method}                & \textbf{Parameter} & \textbf{RoboMimic State} & \textbf{RoboMimic Image} & \textbf{Franka} & \textbf{RoboCrowd} \\ \hline
\multirow{6}{*}{All}           & Optimizer          & \multicolumn{4}{c}{Adam}                                                                   \\
                               & Learning Rate      & \multicolumn{4}{c}{0.0001}                                                                 \\
                               & Batch Size         & \multicolumn{4}{c}{256}                                                                    \\
                               & Training Steps     & 50,000                   & \multicolumn{3}{c}{100,000}                                     \\
                               & Action Chunk       & 1                        & 1                        & 4               & 10                 \\
                               & Image Resolution   & --                       & (84, 84)                 & (128, 128)      & (128, 128)         \\ \hline
\multirow{5}{*}{\abv}           & Augmentations      & --                       & \multicolumn{3}{c}{Random Scale and Crop (0.9, 0.95)}           \\
                               & $\beta$            & 0.05                     & \multicolumn{3}{c}{0.01}                                        \\
                               & Image Recon Weight & --                       & \multicolumn{3}{c}{0.005}                                       \\
                               & $z_s$              & 12                       & 16                       & 24              & 16                 \\
                               & $z_a$              & 6                        & 6                        & 16              & 12                 \\
                               & $k$                & \multicolumn{4}{c}{(5,6,7)}                                                                \\ \hline
\multirow{2}{*}{VIP}           & $z$                & 8                        & 16                       & 24              & 16                 \\
                               & $\gamma$           & \multicolumn{4}{c}{0.98}                                                                   \\ \hline
InfoNCE                        & $z$                & 8                        & 16                       & 24              & 16                 \\ \hline
\multirow{2}{*}{MINE}          & $z$                & 8                        & 16                       & 24              & 16                 \\
                               & $\alpha$           & \multicolumn{4}{c}{0.9}                                                                    \\ \hline
\multirow{4}{*}{Compatibility} & Ensemble Size      & \multicolumn{4}{c}{5}                                                                      \\
                               & Dropout            & \multicolumn{4}{c}{0.5}                                                                    \\
                               & $\eta$             & 0.025                    & 0.05                     & 0.1             & 0.05               \\
                               & $\lambda$           & 8                        & 4                        & 2               & 4                  \\ \hline
\multirow{2}{*}{Uncertainty}   & Ensemble Size      & \multicolumn{4}{c}{5}                                                                      \\
                               & Dropout            & \multicolumn{4}{c}{0.5}                                                                   
\end{tabular}
}
\caption{Hyperparameters for all methods.}
\label{tab:hyperparameters}
\end{table*}
\vspace{0.05in}

\noindent \textbf{Randomized $k$-NN Estimation.} We estimate the mutual information using random batches for $k$-NN estimators. When doing so, we use a batch size of 1024 and iterate over the entire dataset 4 times.
\vspace{0.05in}

\noindent \textbf{Checkpoint Selection.} We train all state-based models for 50K timesteps and all image-based models for 100K timesteps. For VAEs, BC policies, and VIP we select final checkpoints (e.g. 50K or 100K steps). We found that InfoNCE and MINE tended to overfit quite fast. For InfoNCE we used checkpoints after 20K for state and 40K for images. For MINE we used 50K for state and 60K for images.

\subsection{Evaluation Details} 

\noindent \textbf{RoboMimic} For training policies in robomimic, we use the same architecture as in the image-based data quality experiments with an MLP action head using L2 loss. We train for 100K timesteps before running 200 evaluation episodes. Episodes are truncated after 400 timesteps.

\vspace{0.05in}

\noindent \textbf{RoboCrowd.} We use an ALOHA robot setup to evaluate performance on the RoboCrowd benchmark, with ten trials per method. As in \citet{mirchandani2024robocrowd} each trial assigned one of the following scores: 1 point for successfully grasping any number of candies, 2 points for returning any number of candies, and 3 points for returning exactly one candy. 0 points are given otherwise. Policies are trained for 200K timesteps using the same architecture ad hyperparameters from ACT, e.g. encoder-decoder transformer with L1 loss and action chunks of size 100.

\vspace{0.05in}

\noindent \textbf{Franka.} We use the Franka robot setup from DROID \citep{droid}, and run 15 trials per method. We score trails of ``DishRack'' as fully successful (1) if the robot puts both items in the dish rack, and a failure other wise (0). We score trials of ``PenInCup'' as fully successful (1) if the pen ends up completely in the cup a failure in any other scenario (0). We use the same architectures and hyperparameters as in \citet{droid} for evaluation, e.g. Diffusion Policy \citep{chi2023diffusion} with action chunks of size 16 and execution size of 8 with a few differences. Instead of using pre-trained ResNet-50s as in \citet{droid}, we use ResNet-34s initialized from scratch with GroupNorm instead of BatchNorm. To compensate, we train for 200K steps as opposed to 50K.

\end{document}